\documentclass{svproc}
\usepackage{url}
\usepackage{graphicx}

\begin{document}
\mainmatter              
\title{Learning to Extract Cross-Domain Aspects and Understanding Sentiments Using Large Language Models}
\titlerunning{Learning to Extract  Cross-Domain Aspects}  
%
\author{Karukriti Kaushik Ghosh \and Chiranjib Sur}
\authorrunning{Karukriti Ghosh et al.} 
%
\tocauthor{Ivar Ekeland, Roger Temam, Jeffrey Dean, David Grove,
Craig Chambers, Kim B. Bruce, and Elisa Bertino}
\institute{Indian Institute of Technology, Guwahati Assam 781039, India,\\
\email{\{karukriti,chiranjib\}@iitg.ac.in}
}

\maketitle              

\begin{abstract}
Aspect-based sentiment analysis (ASBA) is a refined approach to sentiment analysis that aims to extract and classify sentiments based on specific aspects or features of a product, service, or entity. Unlike traditional sentiment analysis, which assigns a general sentiment score to entire reviews or texts, ABSA focuses on breaking down the text into individual components or aspects (e.g., quality, price, service) and evaluating the sentiment towards each. This allows for a more granular level of understanding of customer opinions, enabling businesses to pinpoint specific areas of strength and improvement. The process involves several key steps, including aspect extraction, sentiment classification, and aspect-level sentiment aggregation for a review paragraph or any other form that the users have provided. ABSA has significant applications in areas such as product reviews, social media monitoring, customer feedback analysis, and market research. By leveraging techniques from natural language processing (NLP) and machine learning, ABSA facilitates the extraction of valuable insights, enabling companies to make data-driven decisions that enhance customer satisfaction and optimize offerings. As ABSA evolves, it holds the potential to greatly improve personalized customer experiences by providing a deeper understanding of sentiment across various product aspects. In this work, we have analyzed the strength of LLMs for a complete cross-domain aspect-based sentiment analysis with the aim of defining the framework for certain products and using it for other similar situations. We argue that it is possible to that at an effectiveness of 92\% accuracy for the Aspect Based Sentiment Analysis dataset of SemEval-2015 Task 12.
\end{abstract}
\section{Introduction}

\begin{figure}[!h]
  \includegraphics[width=\linewidth]{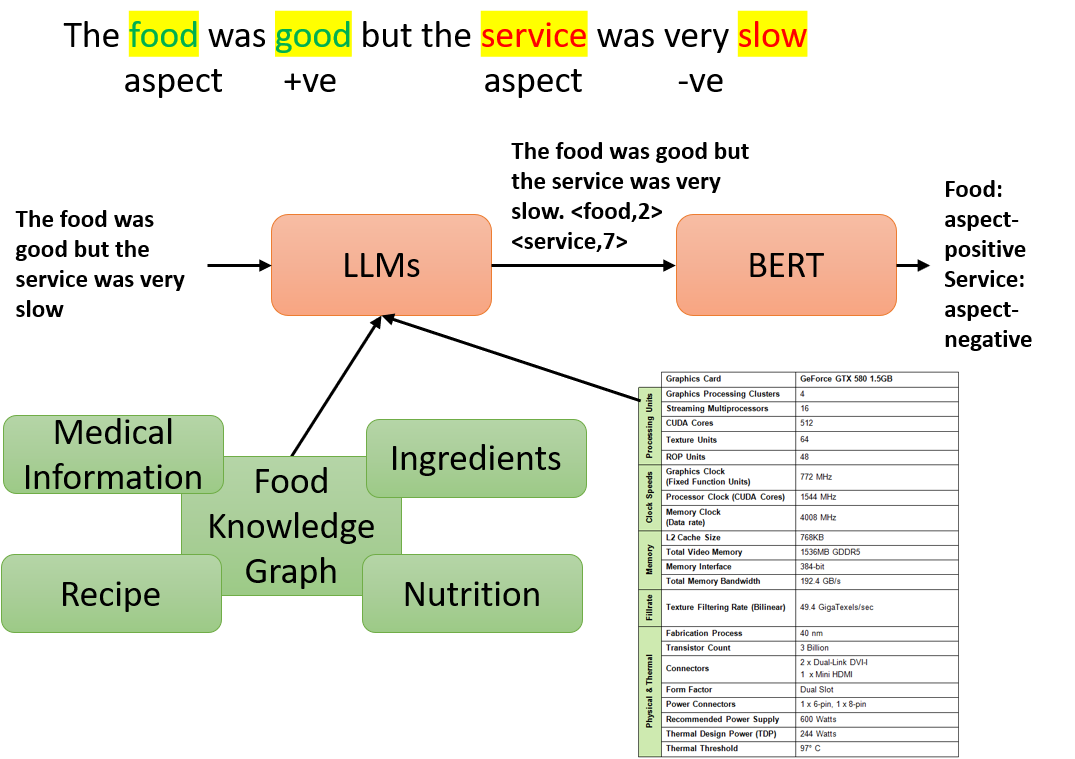}
  \caption{Diagram illustrating how specific knowledge source or knowledge graph combined with Large Language Models can help in understanding the specifics of a product and the sentiments of the reviews can be extracted with more precision.}
  \label{fig:p2_1}
\end{figure}

Aspect-based sentiment analysis (ABSA) \cite{tang2016aspect} is an advanced technique within the broader field of sentiment analysis that goes beyond general sentiment classification. Unlike traditional sentiment analysis, which assigns an overall sentiment (positive, negative, or neutral) to a piece of text, ABSA breaks down the text into specific aspects or features related to the subject being discussed, such as product quality, customer service, pricing, or delivery \cite{jin2020bert}. Each aspect is then evaluated for sentiment, offering a more nuanced and precise understanding of customer opinions \cite{jin2020bert}. The growing reliance on online reviews, social media discussions, and customer feedback has made ABSA an invaluable tool for businesses seeking to understand consumer sentiment at a deeper level \cite{blitzer2007biographies}. By analyzing sentiments tied to individual aspects of a product or service, companies can gain insights into what aspects of their offerings are resonating with customers, which elements require improvement, and how different features contribute to overall satisfaction \cite{blitzer2007biographies}. This allows businesses to make targeted improvements to specific aspects of their products or services rather than making broad changes based on generalized feedback \cite{dodge2020fine}. 
ABSA has applications across various domains, including e-commerce, hospitality, healthcare, and social media monitoring, where understanding the sentiment surrounding individual aspects of a product or service can lead to more informed decision-making \cite{elsayed2018large}. Leveraging natural language processing (NLP) and machine learning techniques, ABSA enables automated and scalable \cite{khosla2020supervised} sentiment analysis, helping businesses process large volumes of data quickly and efficiently \cite{elsayed2018large}. As ABSA continues to evolve, it holds the potential to significantly enhance personalized customer experiences and improve customer satisfaction by providing businesses with actionable, aspect-level insights \cite{ganin2016domain},\cite{gunel2020supervised}, \cite{johnson2018training}.

The overall structure of this document is organized as Section 1 with introduction and literature review, Section 2 has the problem statement, Section 3 has the architectural details, Section 4 has a brief methodology, Section 5 showcases the experimental setup and results, and Section 6 details the conclusion and future work. 

The main contribution of this work: 1) We proposed a framework for cross-domain aspect-based sentiment analysis 2) We modified the existing Aspect Based Sentiment Analysis dataset of SemEval-2015 Task 12 so that we can adopt it for cross-domain sentiment analysis through the use of external knowledge sources and data leverages. 3) We proposed a model that will be more business-friendly as this will require less training and expertise in data science skills 4) We demonstrated that cross-domain does not require finetuning, which is costly mainly to fine-tune a big model like BERT or other Large Language Models.

\section{Problem Statement}
In the age of digital transformation, businesses are increasingly relying on customer feedback \cite{kenton2019bert} from various online platforms such as product reviews, social media, and customer support forums to gauge public perception and improve their products or services. While traditional sentiment analysis methods can provide a broad understanding of whether customers are generally satisfied or dissatisfied, they fail to capture the nuanced opinions expressed about specific aspects \cite{lewis2019bart} or features of a product or service. This presents a significant challenge, as businesses need detailed insights into which particular elements of their offerings are positively or negatively perceived by customers. 
The problem lies in the need for a more granular approach to sentiment analysis—Aspect-Based Sentiment Analysis (ABSA). ABSA allows organizations to extract, identify, and classify sentiments regarding specific aspects or features of a product, service, or entity. For example, in an e-commerce setting, customers may express satisfaction with the quality of a product but dissatisfaction with its price or delivery time. Traditional sentiment analysis would label the review as positive or negative overall, failing to highlight these critical distinctions \cite{lewis2019bart}. 
The key challenge in ABSA is to accurately identify relevant aspects within textual data and classify sentiments towards these aspects in a manner that reflects the context of the review or feedback. This requires advanced techniques in natural language processing (NLP) \cite{khandelwal2020nearest} and machine learning, which can address issues such as ambiguity, aspect extraction, and sentiment classification. Additionally, the ability to scale ABSA models to handle large volumes of diverse data while maintaining high accuracy is a significant hurdle \cite{khandelwal2020nearest}.

By overcoming these challenges, cross-domain ABSA models can provide businesses with a deeper, more actionable understanding of customer opinions and focus on improving them. It can help businesses identify strengths and weaknesses in specific areas, such as product features, customer service, or pricing, leading to more targeted improvements and a better alignment of offerings with customer expectations. Furthermore, ABSA can enhance personalized marketing strategies, improve customer satisfaction, and foster greater customer loyalty by addressing the issues that matter most to individual customers.
Therefore, the problem at hand is to develop an effective and scalable Aspect-Based Sentiment Analysis system that can accurately extract and assess sentiment at the aspect level, providing businesses with the insights they need to optimize their products, services, and customer experiences.

\section{Architecture}
The first part of the architecture deals with understanding what an aspect is and how it varies from product to product and domain to domain. To counter that problem, we have used LLMs (like llama) with external information for better understanding. We can use any model like BERT or trainable parameters for inference. We have trained \cite{khosla2020supervised} the BERT with one set of data (either Laptop or Restaurant) and then tested with both  Laptop and Restaurant to establish the capability for a cross-domain analysis, unlike the other previous works \cite{tang2016aspect}, where they have fine-trained the models for effectiveness. Fine-tuning models can be very costly, mainly when data collection, annotation, and requirements to acquire high-end graphics processing units are concerned. Figure \ref{fig:p2_1} provides the details of the architecture used for experiments. 

The architecture of BERT (Bidirectional Encoder Representations from Transformers) \cite{lewis2019bart} is based on the Transformer model, which consists of an encoder stack responsible for understanding the relationships between words in a sentence. BERT utilizes bidirectional self-attention is based on the following equation $ A(Q,K,V) = \frac{QK^T}{\sqrt{d_k}}V $, allowing the model to consider the context from both the left and right sides of a token during training. Unlike previous models like GPT, which process text in a left-to-right or right-to-left manner, BERT processes the entire sequence of text simultaneously, capturing more complete contextual relationships. This bidirectional approach significantly improves the model's ability to understand nuanced meanings in text, such as sentiment, aspect extraction, and other linguistic patterns. 
The core architecture of BERT consists of multiple layers of Transformer encoders, typically 12 layers in the base model and 24 layers in the large version. Each encoder layer includes two main components: self-attention and feed-forward neural networks. The self-attention mechanism enables the model to weigh the importance of each word in the sequence relative to all other words, capturing dependencies between words regardless of their position in the sentence. The feed-forward neural network processes these weighted embeddings further to refine the contextual understanding. BERT uses position embeddings to account for the sequential order of tokens, ensuring that the model understands word positions even in a bidirectional setting. Additionally, token embeddings and segment embeddings are added to the input tokens to represent the meaning of each word and differentiate between different parts of the input, such as the aspect and the review text in Aspect-Based Sentiment Analysis (ABSA). The output of BERT is a set of contextualized token embeddings, with each token in the sequence having its own learned representation. For tasks like ABSA, the output representation of the [CLS] token, placed at the beginning of the input, is typically used for classification tasks (e.g., sentiment classification), while the embeddings of individual tokens can be used for aspect extraction. 
Finally, BERT is pre-trained on vast amounts of text data in an unsupervised manner and can be fine-tuned for specific downstream tasks. In the case of ABSA, it can be fine-tuned to identify aspects in text and predict sentiment scores related to each aspect.

\begin{figure}[!h]
  \includegraphics[width=\linewidth]{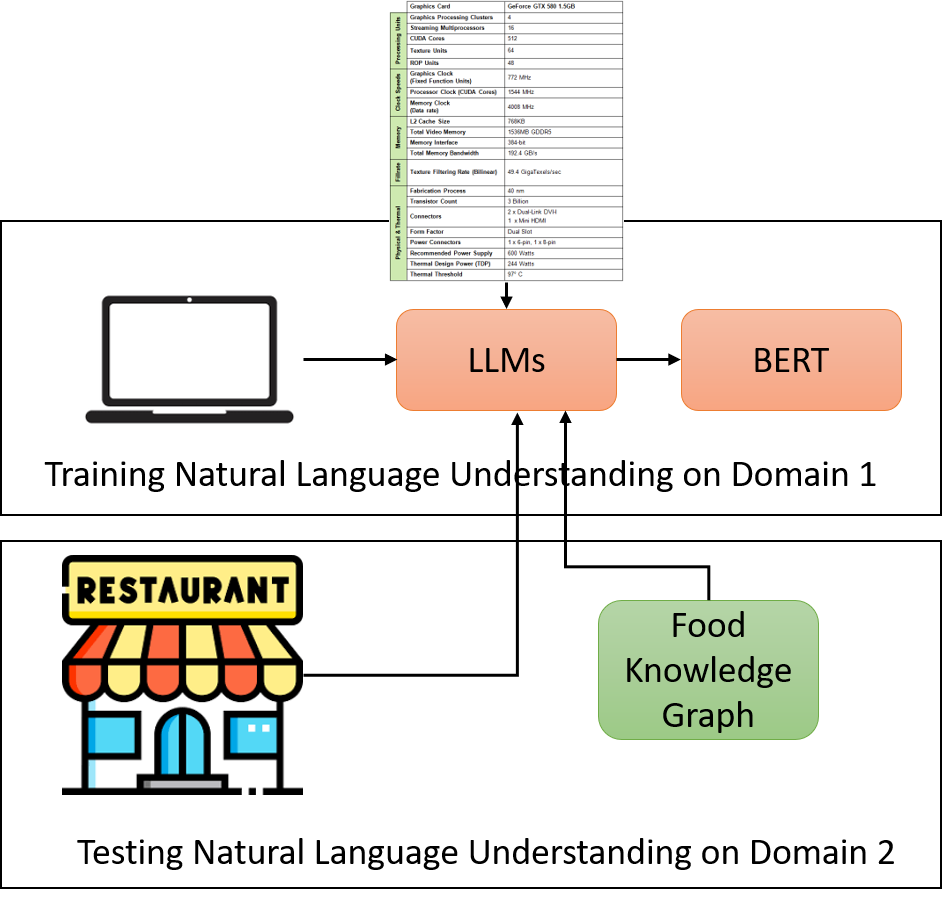}
  \caption{Diagram illustrating the Domain Adaptation Experiment for Aspect-Based Sentiment Analysis (ABSA). It shows the process of training the ABSA model on Domain A (e.g., restaurant reviews) and then adapting it to Domain B (e.g., laptop reviews) without further fine-tuning. The diagram emphasizes the transfer of learned knowledge, with components such as aspect extraction, sentiment classification, and domain-specific adaptation clearly labeled. This approach helps in evaluating how well the model performs when transferred to a different domain with varying aspects and sentiment patterns.}
  \label{fig:fig:p2_2}
\end{figure}

\section{Methodology}
The methodology for Aspect-Based Sentiment Analysis (ABSA) is designed to address the key steps involved in extracting and analyzing sentiment at the aspect level of customer feedback. This process begins with aspect extraction, where the goal is to identify relevant product or service features mentioned within the text. Various methods can be applied to this step, including deep memory network and pure BERT fine-tuning approaches, where keywords and phrases related to aspects (such as "battery life," "screen quality," "customer service") are used to locate aspects within a document. Additionally, machine learning techniques were employed in many of the previous works like \cite{tang2016aspect}, such as supervised models trained on labeled data, which are using part-of-speech tagging and dependency parsing to automatically identify aspects. Unsupervised techniques, such as topic modeling (e.g., Latent Dirichlet Allocation or LDA), can also be used to identify aspects by clustering semantically related terms, but the fundamental problem of better accuracy still remian persistent. 

It has been observed that once the aspects are identified, the classification can be very easy, and hence, some traditional models train joint decision models. However, in this work, we have first dealt with aspect detection, and in the last step, we have done sentiment classification, which involves assigning a sentiment score (positive, negative, or neutral) to each extracted aspect. Several sentiment classification methods can be employed, ranging from traditional machine learning classifiers such as Naive Bayes, Support Vector Machines (SVM), or Logistic Regression to deep learning techniques like Recurrent Neural Networks (RNNs) or Long Short-Term Memory networks (LSTMs). Since, businesses are prone to decision making with risk, advanced models like BERT and LLMs will be more trustworthy. These models are trained on labeled data, where each aspect is associated with a sentiment label, allowing the model to learn how to classify sentiment accurately. In the case of deep learning and Large Language Models, pre-trained models like BERT (Bidirectional Encoder Representations from Transformers) can be either trained from scratch or fine-tuned for the specific task of sentiment classification, leveraging large-scale semantic knowledge embedded in the model.  The sentiment analysis model is designed to handle both local context (the sentiment towards a specific aspect) and global context (overall sentiment towards the product or service). To avoid training for weeks on huge computing resources, we have adopted the 12 layered BERT model for our experiments. The experiments were done on Nvidia 3090 RTX with 24 GB RAM installed in a workstation with 64 GB RAM and Intel 13900K CPU. 

The final stage in ABSA involves evaluating the performance of the system. We have used accuracy in this work following the previous works like \cite{tang2016aspect}. Since the dataset is well-balanced, accuracy is used to assess the effectiveness of both aspect extraction and sentiment classification. It would have been great if we can use metrics to help in measuring how well the system can identify and classify aspects, as well as how accurately it can assess the sentiment associated with each aspect. However, since the  Aspect Based Sentiment Analysis dataset of SemEval-2015 Task 12, has a particular way of representing the joint (both aspect and sentiment) ground truth, we have adopted that. Furthermore, domain adaptation plays an essential role in ensuring that ABSA models are robust and transferable across different industries and product categories. Training models on domain-specific data and testing them across various contexts—such as electronics, fashion, or hospitality—ensures that the model generalizes well and performs optimally in multiple settings. Techniques like transfer learning was avoided through the use of LLMs like llama. Where the last model is trained once on one domain dataset and is not fine-tuned for another, which was crucial for enhancing model versatility in many previous works and minimizing the need for large labeled datasets in every domain.

Finally, bias detection and fairness are integral parts of any model training and testing methodology, especially in ensuring that the sentiment analysis does not disproportionately favor or penalize certain demographic groups or product features. To identify potential biases, whether due to unequal data representation or biased labeling, we have investigated the models through masking some of the critical works during training. This is important to ensure fairness and accuracy in the results in case the review text has errors or mistakes. Techniques like re-weighting training samples, data augmentation, and bias mitigation algorithms were also used to increase the dataset size and this helped address the challenge of bias for one category, ensuring that ABSA results reflect diverse customer perspectives without unjust bias. 
This methodology provides a comprehensive framework for implementing Aspect-Based Sentiment Analysis, enabling businesses to extract detailed insights from customer feedback, improve decision-making, and refine their products and services based on specific aspects that matter most to their customers.

\section{Experimentation and Results} 
To effectively implement and evaluate Aspect-Based Sentiment Analysis (ABSA), several experiments has been designed to test various components of the ABSA pipeline, including aspect extraction, sentiment classification, and overall system performance. The following experiments aim to address key challenges in ABSA, evaluate different techniques, and optimize model accuracy and scalability. The first task is to make the model independent of the multiple training phases and to evaluate the performance of an integrated system that combines both aspect extraction and sentiment classification. We have used a pre-trained BERT model. However, any other transformer-based models can be used as long as they are trained to understand the language of aspect extraction and sentiment classification. Then we can fine-tune the model on labeled data once (e.g., product reviews with aspects and sentiments annotated) to improve performance and then free to use it for many such applications. We have used them for only one dataset as the Aspect Based Sentiment Analysis dataset of SemEval-2015 Task 12, has only two products, that is Laptop and Restaurant. 
Testing the model across different domains (e.g., electronics, fashion, food) could have been fruitful to evaluate its generalizability and will surely be a future work. However, since the datasets are from diverse fields, it is high likely to generalize well. Accuracy and F1-score for both aspect extraction and sentiment classification tasks are used. Since the dataset is well balanced, there are hardly any variations between Accuracy and F1-score for all our experiments and the trend of the results went to the same direction. The percentage of correctly extracted aspects is desired but due to the annotation of the data and to compare with other similar works, we have used sentiment-labeled aspects as a benchmark. Evaluation of the model's performance in real-time processing, especially when handling large volumes of text data can be critical and that is why we have considered 12 layered BERT model. This experiments aim to assess how well a single integrated model can handle both aspects of ABSA and whether deep learning techniques like BERT can provide a significant improvement over traditional methods.

Another important of our experiments is to prove domain adaptation and we have also performed experiment to prove that. To evaluate the performance of ABSA models, we have applied the model to different domains (e.g., Laptop when trained with Restaurant, and vice versa). For BERT model, the previous perception was that if we trained on one domain (e.g., Laptop), it may not perform well on another (e.g., Restaurant) due to the difference in language, terminology, and aspect categories. However, we have proved that it can be neutralized, if use LLMs with external source and then use that information to train the BERT model on a domain-specific dataset (e.g., reviews for smartphones) and test the model's performance on another domain (e.g., restaurant reviews). We compare the performance of a domain-specific model versus a generalized model trained across multiple domains. 
We have reported both the Inter-domain and Cross-domain Performance and accuracy of the BERT model is reported in this work. In spite of the absence of transfer learning the BERT pre-trained models (e.g., on general datasets) performed well and the performance remain less varying when fine-tuned on domain-specific datasets.
These experiment should help identify whether ABSA models require domain-specific training or if a generalized model can be adapted to perform well across various industries with minimal fine-tuning. 

\begin{table}
\caption{The Results for Different Experiments with Variation in Models and Cross-Domain Strategy. The results are compared with the results for Deep Memory Network in \cite{tang2016aspect}. }
\begin{center}
\begin{tabular}{r@{\quad}cc}
\hline
\multicolumn{1}{l}{\rule{0pt}{12pt}
                   Architecture and Details}&\multicolumn{2}{l}{Accuracy}\\[2pt]
\hline\rule{0pt}{12pt}
\textbf{Deep Memory Network} \cite{tang2016aspect}  &  & \\
For Laptop & 72.21 & \\
For Restaurant & 80.95 & \\
\textbf{Normal 12 Layer BERT Fine-Tuned}   &     & \\
For Laptop & 82.3 & \\
For Restaurant & 81.5 & \\
\textbf{LLMs for Aspect and 12 Layer BERT}   &     & \\
For Laptop (BERT Trained with Laptop) & 92.1 & \\
For Restaurant  (BERT Trained with Laptop) & 88.9 & \\
\textbf{LLMs for Aspect and 12 Layer BERT}   &     & \\
For Laptop  (BERT Trained with Restaurant) & 90.4 & \\
For Restaurant  (BERT Trained with Restaurant) & 91.4 & \\
\textbf{LLMs for Aspect and 12 Layer BERT}   &     & \\
For Laptop  (BERT Trained with Laptop and Restaurant) & 91.1 & \\
For Restaurant  (BERT Trained with Laptop and Restaurant) & 90.6 & \\[2pt]
\hline
\end{tabular} \label{table1}
\end{center}
\end{table}

From Table \ref{table1}, we can clearly see that the performance of the cross-domain aspects improved significantly without the requirement to fine-tune the BERT models. We have also demonstrated that with the introduction of Large Language Models (like llama), we can make the BERT model much more powerful and easy to learn the aspects and their surrounding sentiments. Additionally, we have shown that the model can become much more versatile and will require less fine-tuning. This is because of the fact that the LLMs can intermediate and help in bridging the gap in the terminology and understanding of what is meant to be domain knowledge. Thus it can be much less work for business people and can adopt the model pipeline without the need for experts and further annotations. 

\section{Conclusion and Future Works}
Aspect-Based Sentiment Analysis (ABSA) represents a significant advancement in the field of sentiment analysis by offering businesses the ability to gain a more nuanced and detailed understanding of customer opinions. Unlike traditional sentiment analysis, which only provides a broad sentiment label, ABSA breaks down the feedback into specific aspects, such as product quality, customer service, pricing, or delivery time, and assesses the sentiment associated with each. This granular level of analysis allows businesses to pinpoint exactly which aspects of their products or services are being praised or criticized, enabling them to make more targeted improvements and drive greater customer satisfaction. 
In this work, we have demonstrated we can use the external knowledge from LLMs and other information, and can improve the performance of Aspect-Based Sentiment Analysis models and with the help of formulation of a cross-domain adaptation, we can set free the need for further training and debugging.

The ability to leverage ABSA can lead to more informed decision-making, as it allows companies to identify key drivers of customer satisfaction or dissatisfaction. Furthermore, ABSA enables the customization of marketing strategies, customer service improvements, and product development based on detailed customer feedback. Cross-domain ABSA facilitates the automation of sentiment analysis across large volumes of data, helping businesses keep pace with the growing amount of customer-generated content on digital platforms. 
Despite its advantages, ABSA still faces challenges such as accurate aspect extraction, dealing with ambiguous language, and maintaining scalability across diverse data sets. However, ongoing advancements in NLP and machine learning continue to improve the accuracy and efficiency of ABSA models, making them an invaluable tool for businesses aiming to stay competitive in today’s data-driven landscape. In conclusion, ABSA not only enhances the way businesses understand customer sentiment but also provides actionable insights that can directly inform product development, customer service, and overall business strategy. As the field continues to evolve, ABSA is poised to play a critical role in helping businesses deliver more personalized, customer-centric experiences that drive growth and customer loyalty.

%
%

\end{document}